%% file: main.tex
\RequirePackage[loading]{tracefnt}

\documentclass[letterpaper,10pt,conference]{ieeeconf}

\IEEEoverridecommandlockouts
\usepackage{docmute}
\input{settings}

\newcommand{\engtitle}{An Efficient Accelerator for Deep Learning-based Point Cloud Registration on FPGAs}
\title{\LARGE \bf \engtitle}

\author{
    Keisuke Sugiura$^{1}$ and Hiroki Matsutani$^{1}$
    \thanks{$^{1}$The authors are with Keio University, 3-14-1, Hiyoshi, Kohoku, Yokohama, Kanagawa 223-8522, Japan. \texttt{sugiura@arc.ics.keio.ac.jp}, \texttt{matutani@arc.ics.keio.ac.jp}}
}

\hypersetup{
  pdftitle={\engtitle},
  pdfauthor={Keisuke Sugiura, Hiroki Matsutani}
}

\begin{document}

\maketitle
\thispagestyle{empty}
\pagestyle{empty}

% Abstract and keywords
\input{abst}

% Header

% Body
\input{intro}

\input{related}

\input{preliminaries}

\input{design}

\input{implementation}

\input{eval}

\input{conc}

% Bibliography
% \renewcommand{\baselinestretch}{1.0}
\bibliographystyle{IEEEtran}

\input{refer.tex}
\end{document}

%% file: settings.tex
\usepackage{amsmath}
\usepackage{amssymb}
\usepackage{amsfonts}
\usepackage{bm}
\usepackage{comment}
\usepackage{fancybox}
\usepackage{framed}
\usepackage{color}
\usepackage{graphicx}
\usepackage{multicol}
\usepackage{multirow}
\usepackage{cite}
\usepackage{hyperref}
\hypersetup{
    colorlinks = true,
    urlcolor = blue,
    linkcolor = red,
    citecolor = green
}

\usepackage{algorithm}
\usepackage[noend]{algpseudocode}
\usepackage{algorithmicx}

\renewcommand{\baselinestretch}{0.96}

\expandafter\def\expandafter\UrlBreaks\expandafter{\UrlBreaks
  \do\a\do\b\do\c\do\d\do\e\do\f\do\g\do\h\do\i\do\j%
  \do\k\do\l\do\m\do\n\do\o\do\p\do\q\do\r\do\s\do\t%
  \do\u\do\v\do\w\do\x\do\y\do\z\do\A\do\B\do\C\do\D%
  \do\E\do\F\do\G\do\H\do\I\do\J\do\K\do\L\do\M\do\N%
  \do\O\do\P\do\Q\do\R\do\S\do\T\do\U\do\V\do\W\do\X%
  \do\Y\do\Z}

\DeclareMathOperator*{\argmin}{arg\,min}

\DeclareMathOperator{\SE}{\mathrm{SE}(3)}

\algrenewcommand\algorithmicindent{1.0em}

\algnewcommand\algorithmicforeach{\textbf{for each}}
\algdef{S}[FOR]{ForEach}[1]{\algorithmicforeach\ #1\ \algorithmicdo}

\algnewcommand\AlgAnd{\textbf{and} }
\algnewcommand\AlgOr{\textbf{or} }
\algnewcommand\AlgContinue{\textbf{Continue}}

\algrenewcommand\textproc{}

\algnewcommand{\Initialize}[1]{
	\State \textbf{Initialize:}
 	\State \hspace*{\algorithmicindent}\parbox[t]{0.8\linewidth}{\raggedright #1}}

\algnewcommand{\LeftComment}[1]{
    \Statex $\triangleright$ #1 \hfill}

\def\BibTeX{{\rm B\kern-.05em{\sc i\kern-.025em b}\kern-.08em
    T\kern-.1667em\lower.7ex\hbox{E}\kern-.125emX}}

\usepackage{CJKutf8}

%% file: abst.tex
% abst.tex

\begin{abstract}
% Point cloud registration enables the 3D environment perception and many robotic applications such as odometry and SLAM (Simultaneous Localization And Mapping).
% These applications are increasingly important for autonomous mobile robots with limited computational resources, thereby motivating the development of resource-efficient registration method on low-power FPGAs.
Point cloud registration is the basis for many robotic applications such as odometry and Simultaneous Localization And Mapping (SLAM), which are increasingly important for autonomous mobile robots.
Computational resources and power budgets are limited on these robots, thereby motivating the development of resource-efficient registration method on low-cost FPGAs.
% In this paper, we propose a novel approach for FPGA-based 3D point cloud registration.
% Especially, we focus on a recent deep learning-based method, PointNetLK, considering its simplicity, scalability, and resource efficiency.
In this paper, we propose a novel approach for FPGA-based 3D point cloud registration built upon a recent deep learning-based method, PointNetLK.
A highly-efficient FPGA accelerator for PointNet-based feature extraction is designed and implemented on both low-cost and mid-range FPGAs (Avnet Ultra96v2 and Xilinx ZCU104).
% We conduct design optimizations by taking full advantage of the inherent parallelism in PointNet.
Our accelerator design is evaluated in terms of registration speed, accuracy, resource usage, and power consumption.
Experimental results show that PointNetLK with our accelerator achieves up to 21.34x and 69.60x faster registration speed than the CPU counterpart and ICP, respectively, while only consuming 722mW and maintaining the same level of accuracy.
\end{abstract}

%% file: intro.tex
% intro.tex

\section{Introduction} \label{sec:intro}
Point cloud registration is the process of estimating a rigid transform that best aligns a pair of point clouds.
It is the key component for 3D reconstruction and robotic applications, e.g., odometry and SLAM, which are increasingly important in autonomous mobile robots with limited computational resources and power budgets.
This necessitates the implementation of a lightweight registration method running on low-power mobile devices such as FPGA SoCs.

Inspired by the tremendous success of deep learning, significant advancements have been made in the development of learning-based registration methods over the past few years.
PointNetLK~\cite{Aoki19} is a representative learning-based method, which combines the Lucas-Kanade (LK)-based pose estimation and PointNet feature embedding.
The conventional geometry-based methods such as ICP~\cite{Besl92} and many other learning-based methods rely on point correspondences to obtain the rigid transform in closed form, which results in the computational complexity of $O(N^2)$, where $N$ is the number of points.
In contrast, PointNetLK has $O(N)$ complexity, and PointNet is a quite small and easy-to-implement neural network.
Combined with these, PointNetLK brings a better scalability and a certain performance advantage; it is therefore suitable for resource-limited computing platforms.

From these considerations, in this paper, we propose a highly-efficient design of PointNetLK targeting resource-limited FPGA SoCs.
As PointNet feature extraction becomes a performance bottleneck, we develop a dedicated accelerator IP core for PointNet, and implement it on the FPGA logic circuit.
We use Xilinx ZCU104 Evaluation Kit as an affordable mid-range FPGA SoC.
We fully optimize the IP core design by leveraging a high degree of parallelism in PointNet.
% The memory footprint of the resulting IP core is $O(1)$.
Experiments demonstrate that our accelerator improves the performance by a large margin without degrading the generalization ability and accuracy.
% We also show that it provides faster registration speed than ICP and the same level of accuracy.
We conduct weight quantization to further reduce the resource usage, and show that the quantized IP core can be implemented on even smaller and low-cost FPGAs (Avnet Ultra96v2).

The rest of the paper is organized as follows: Section \ref{sec:related} overviews related works.
Section \ref{sec:background} formulates the registration problem and describes PointNetLK algorithm.
Section \ref{sec:design} illustrates design optimizations carried out for our FPGA SoC-based implementation.
Evaluation results in terms of speed, accuracy, resource utilization, and power consumption are presented in Section \ref{sec:eval}.
Section \ref{sec:conc} concludes the paper.

%% file: related.tex
% related.tex

\section{Related Works} \label{sec:related}
\subsection{Deep Learning-based Point Cloud Registration} \label{sec:related-dnn-registration}
Deep learning techniques have been successfully applied to a registration problem, outperforming conventional geometry-based methods such as ICP and its variants~\cite{Besl92,Segal09}.
There exists a line of work that employs DNNs to predict a rigid transform from input point sets in an end-to-end fashion~\cite{JiaxinLi17,Valente19,LiDing19,Sarode19,Pais20}.
Another approach combines DNN feature extraction and non-learning-based closed-form pose estimation.
LORAX~\cite{Elbaz17} leverages a shallow autoencoder to extract a feature descriptor from a subset of points, and computes a rigid motion from the matched descriptor pairs.
A number of methods \cite{YueWang19A,WeixinLu19,YueWang19B,ZiJianYew20,Choy20,Kurobe20,KexueFu21,TaewonMin21} predict point correspondences using DNNs, and perform SVD to compute a rigid transform.
Aside from these methods, other learning-based methods perform the iterative registration in the framework of the LK algorithm.
PointNetLK~\cite{Aoki19} is a representative method; it iteratively refines a rigid transform by aligning global point cloud features extracted by PointNet.
Li \textit{et al.}~\cite{XueqianLi21} improves the generalization ability of PointNetLK by analytically computing a Jacobian instead of approximating it.

\subsection{FPGA-based Acceleration for Point Cloud Registration} \label{sec:related-fpga-registration}
Despite the growing importance, only a few works have investigated the FPGA-based acceleration of 3D point registration.
Kosuge \textit{et al.}~\cite{Kosuge20} develop an accelerator for the ICP-based object pose estimation, which is a critical process in picking robots.
They focus on the $k$-nearest neighbor ($k$-NN) search, which constitutes a major bottleneck in ICP, and devise a novel hierarchical graph data structure for improved efficiency.
The proposed accelerator combines a parallelized distance computation unit and a dedicated sorter unit to speed up the graph construction and NN search.
Deng \textit{et al.}~\cite{QiDeng21} present an FPGA-based accelerator for Normal Distributions Transform (NDT).
NDT~\cite{Biber03} models point clouds as a set of voxels, each of which represents the Gaussian distribution of points.
They introduce a new hierarchical memory-efficient data structure to accelerate the voxel search operations.
Eisoldt \textit{et al.}~\cite{Eisoldt21} implement Truncated Signed Distance Function (TSDF)-based registration method and TSDF map update process onto the FPGA logic circuit for efficient 3D TSDF-based LiDAR SLAM.
These works successfully demonstrate the effectiveness of FPGA acceleration for well-established and geometry-based registration methods.
This paper is the first to explore the FPGA-based accelerator design for a learning-based method.

%% file: preliminaries.tex
\section{Background} \label{sec:background}
\subsection{PointNetLK Algorithm} \label{sec:background-pointlk}
In this section, we briefly describe the PointNetLK algorithm.
We refer to \cite{Aoki19,XueqianLi21} for the detailed derivation.

% \subsection{3D Point Registration} \label{sec:background-point-registration}
The aim of point registration is to align two 3D point clouds, referred to as a \textbf{template} $\mathcal{P}_T$ and \textbf{source} $\mathcal{P}_S$, by estimating a 3D rigid transform $\mathbf{G} \in \SE$ from $\mathcal{P}_S$ to $\mathcal{P}_T$.
PointNetLK finds an optimal transform such that global features of two point clouds are equal: $\bm{\phi}(\mathbf{G} \cdot \mathcal{P}_S) = \bm{\phi}(\mathcal{P}_T)$.
$\bm{\phi}: \mathbb{R}^{N \times 3} \to \mathbb{R}^K$ denotes a PointNet that maps a point cloud of $N$ points into a $K$-D global feature.
The transform $\mathbf{G}(\bm{\xi}) = \exp(\bm{\xi}^\wedge)$ is computed from a 6D twist $\bm{\xi} \in \mathbb{R}^6$ via the exponential map.
The definition of wedge operator ($^\wedge$) is found in \cite{Barfoot17}.
For efficiency, PointNetLK swaps the roles of template and source; it computes a twist $\bm{\xi}$ such that the rigid transform $\mathbf{G}(\bm{\xi})^{-1} = \exp(-\bm{\xi}^\wedge)$ from $\mathcal{P}_T$ to $\mathcal{P}_S$ minimizes the difference between $\bm{\phi}(\mathcal{P}_S)$ and $\bm{\phi}(\mathbf{G}(\bm{\xi})^{-1} \cdot \mathcal{P}_T)$:
\begin{equation}
  \bm{\xi}^* = \textstyle \argmin_{\bm{\xi}} \left| \bm{\phi}(\mathcal{P}_S)
    - \bm{\phi}(\mathbf{G}(\bm{\xi})^{-1} \cdot \mathcal{P}_T) \right|^2.
    \label{eq:pointlk-problem}
\end{equation}
By applying the first-order Taylor expansion, we linearize the term $\bm{\phi}(\mathbf{G}(\bm{\xi})^{-1} \cdot \mathcal{P}_T)$:
\begin{equation}
  \bm{\phi}(\mathbf{G}(\bm{\xi})^{-1} \cdot \mathcal{P}_T)
    \simeq \bm{\phi}(\mathcal{P}_T) + \mathbf{J} \bm{\xi}, \label{eq:pointlk-linearize}
\end{equation}
where $\mathbf{J} = \frac{\partial}{\partial \bm{\xi}} \bm{\phi}(\mathbf{G}(\bm{\xi})^{-1} \cdot \mathcal{P}_T) \in \mathbb{R}^{K \times 6}$ is a Jacobian matrix.
% \begin{equation}
%   \mathbf{J} = \frac{\partial}{\partial \bm{\xi}}
%     \bm{\phi}(\mathbf{G}(\bm{\xi})^{-1} \cdot \mathcal{P}_T).
%     \label{eq:pointlk-jacobian}
% \end{equation}
Each column vector $\mathbf{J}_i \in \mathbb{R}^K$ of $\mathbf{J}$ is computed by numerical gradient approximation as follows:
\begin{equation}
  \mathbf{J}_i \simeq t_i^{-1} \left( \bm{\phi}(\exp(-t_i \bm{e}_i) \cdot \mathcal{P}_T)
    - \bm{\phi}(\mathcal{P}_T) \right), \label{eq:pointlk-jacobian2}
\end{equation}
where $t_i$ is an infinitesimal perturbation to the twist and $\bm{e}_i \in \mathbb{R}^6$ is a unit vector whose $i$-th element is 1 and the others are 0.
It turns out that $\mathbf{J}$ is computed only once at the initialization.
By substituting Eq. \ref{eq:pointlk-linearize} into Eq. \ref{eq:pointlk-problem}, we can solve for the optimal twist $\bm{\xi}$ as follows:
\begin{equation}
  \bm{\xi} = \mathbf{J}^\dagger \left( \bm{\phi}(\mathcal{P}_S)
    - \bm{\phi}(\mathcal{P}_T) \right), \label{eq:pointlk-solution}
\end{equation}
where $\mathbf{J}^\dagger$ is a pseudo-inverse of $\mathbf{J}$.
We transform the source $\mathcal{P}_S$ using $\mathbf{\Delta G} = \exp \left( \bm{\xi}^\wedge \right)$ and proceed to the next iteration until convergence.
The final solution $\mathbf{G}$ is obtained as a product of all incremental transforms, i.e., $\mathbf{G} = \mathbf{\Delta G}_n \cdots \mathbf{\Delta G}_2 \mathbf{\Delta G}_1$, where $\mathbf{\Delta G}_k$ is the estimate at the $k$-th iteration, and $n$ is the number of iterations.

\subsection{Advantages of PointNetLK} \label{sec:background-pointlk-advantages}
PointNet is a simple yet powerful network for point cloud processing, which contributes to the computational efficiency, low memory consumption, and ease of implementation.
The network consists of five fully-connected layers (see Fig. \ref{fig:pointnet-layers}), each of which is followed by batch normalization and ReLU activation.
These are responsible for extracting a $K$-D point-wise local feature from a given point.
The max-pooling layer is placed at the end to aggregate point-wise local features and compute a global feature.
The preprocessing such as normal estimation are not required, as it directly processes raw 3D point coordinates.

PointNetLK does not depend on the correspondences, and is instead based on the alignment of global point cloud features.
Importantly, PointNet has a computational and space complexity of $O(N)$, so does the PointNetLK.
This leads to a significant advantage compared to the correspondence-based methods mentioned in Section \ref{sec:related-dnn-registration}, which has $O(N^2)$ complexity due to the correspondence search.
As shown later, the on-chip memory consumption of our PointNet accelerator is $O(1)$, since it processes input points one-by-one and requires a storage for only one $K$-D local feature.
PointNet does not even contain convolutional layers, skip connections, and looping structures.
Fully-connected layers are amenable to massive parallelization on the FPGA circuit.
The data flows in one direction from the input to output layer, thus PointNet is suitable for the inter-layer pipelining.

% As shown in Fig. \ref{fig:pointnet-layers}, PointNet is a simple yet powerful DNN model for point cloud processing, e.g., classification and segmentation.
% The preprocessing and handcrafted feature extraction such as normal estimation are not required, as it directly processes raw 3D point coordinates $(x, y, z)$ as an input and generates a $K$-D global feature $\bm{\phi}(\mathcal{P})$.
% PointNet is basically a stack of fully-connected layers which are highly parallelizable; it does not even contain convolutional layers, skip connections, and looping structures.
% In other words, the data flows in one direction from the input to output layer (analogous to a bucket brigade), making it suitable for the inter-layer pipelining.

%% file: design.tex
% design.tex

\section{Design of PointNet Accelerator} \label{sec:design}
\subsection{Overview of the Design} \label{sec:design-overview}
\begin{figure}[htbp]
  \centering
  \includegraphics[keepaspectratio,width=0.85\linewidth]{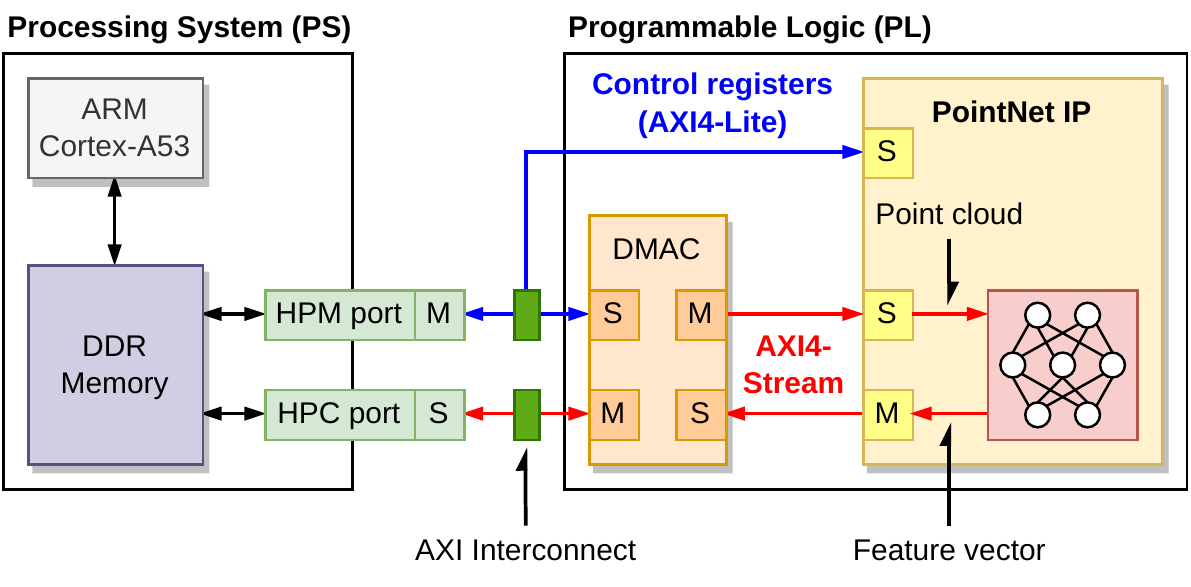}
  \vspace*{-7.5pt}
  \caption{Board-level implementation (Xilinx Zynq UltraScale+ MPSoC)}
  \label{fig:block-diagram}
\end{figure}

This section presents an FPGA SoC-based design of PointNet accelerator, since PointNet feature extraction becomes a major bottleneck in PointNetLK as described in Section \ref{sec:eval}.
Fig. \ref{fig:block-diagram} depicts a block diagram of our board-level implementation, which is partitioned into the processing system (PS) part and the programmable logic (PL) part.
The proposed PointNet IP core and a Direct Memory Access (DMA) controller are instantiated inside the PL part, which computes a global feature from an input point cloud upon a request from the PS part.
The PS part is responsible for setting up the IP core and triggering a DMA controller.
Other steps such as Jacobian computation and coordinate transformation are also performed on the PS part.
For the high-speed data transfer, the DMA controller is connected to a 32-bit wide high-performance slave port (HPC port) and utilizes AXI4-Stream protocol (red lines in Fig. \ref{fig:block-diagram}).
The control registers are accessible through the AXI4-Lite interface connected to a high-performance master port (HPM port, blue lines in Fig. \ref{fig:block-diagram}).

% For ease of explanation, we assume Xilinx Zynq UltraScale+ MPSoC as a target device.

% The DMA transfer between PS--PL is outlined as follows.
% PointNetLK running on the PS part copies input data to the dedicated DRAM buffer with continuous physical addresses.
% It then initiates the DMA transfer by writing a starting physical address and a buffer size to the control registers of a DMA controller.
% Through HPC port, the DMA controller successively reads out the contents on the specified memory space, and passes them to the PointNet core.
% It also receives the PointNet core outputs and writes them back to the DRAM buffer via HPC port.

Our PointNet core has two modes: \textbf{weight initialization} and \textbf{feature extraction}.
In the weight initialization mode, the IP core receives PointNet model parameters (e.g., weight and bias) through the AXI4-Stream interface and stores them to the on-chip BRAM buffer.
The IP core returns a nonzero 32-bit value as an acknowledgement message to notify that the initialization is complete and it is in ready state.
In the feature extraction mode, a 1024D global feature $\bm{\phi}(\mathcal{P}) \in \mathbb{R}^{1024}$ is firstly initialized with zeros.
Then, as illustrated in Fig. \ref{fig:pointnet-layers}, the IP core receives 3D coordinate $\bm{p} \in \mathbb{R}^3$ for each point and computes a 1024D point-wise local feature $\bm{\psi}(\bm{p}) \in \mathbb{R}^{1024}$ by propagating through five consecutive MLP layers.
The global feature $\bm{\phi}(\mathcal{P})$ is updated by taking the element-wise maximum of $\bm{\phi}(\mathcal{P})$ and $\bm{\psi}(\bm{p})$ (Eq. \ref{eq:max-pool}).
In this way, point-wise local features $\bm{\psi}(\bm{p})$ are aggregated into one global feature $\bm{\phi}(\mathcal{P})$.
After the computation is done for all points, the current $\bm{\phi}(\mathcal{P})$ is returned to PS as a final result.
Our design takes advantage from the property of PointNet: the computation for each point is independent except the last max-pooling layer.
This substantially reduces the BRAM consumption, as it obviates the need to keep intermediate results and local features for all points.
The design is also flexible and scalable in a sense that it does not limit the number of input points.
To prevent the accuracy loss, our design uses the 32-bit fixed-point format.

% At the beginning, PS transfers a 32-bit integer (0 or 1) to select the operation mode.

\subsection{Modules in the PointNet IP core} \label{sec:design-modules}
\begin{figure}[htbp]
  \centering
  \includegraphics[keepaspectratio,width=0.55\linewidth]{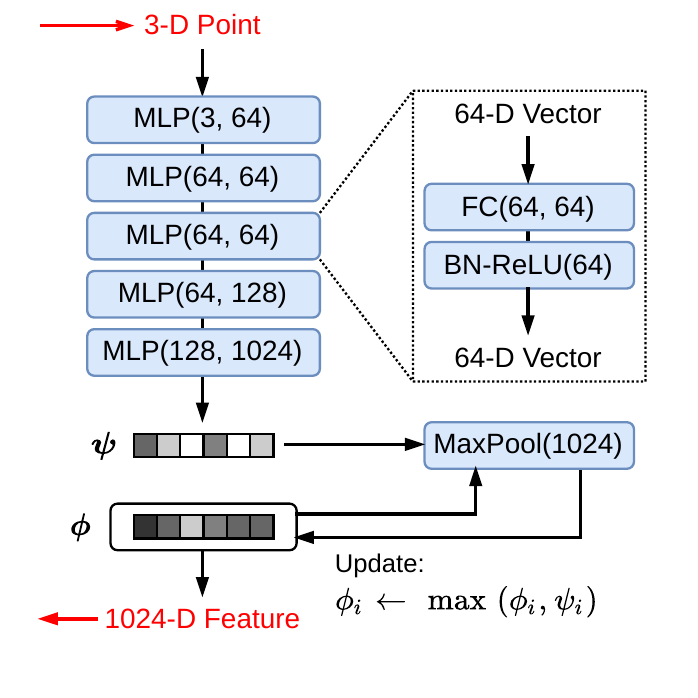}
  \vspace*{-15pt}
  \caption{Computation flow inside the PointNet IP core}
  \label{fig:pointnet-layers}
\end{figure}

As shown in Fig. \ref{fig:pointnet-layers}, the IP core is composed of three types of modules: \textbf{FC}, \textbf{BN-ReLU}, and \textbf{MaxPool}.
\textbf{FC}$(K, L)$ corresponds to a fully-connected layer; it computes a $L$-D output $\bm{y} = \mathbf{W} \bm{x} + \bm{b}$ from a $K$-D input $\bm{x} \in \mathbb{R}^K$, where $\mathbf{W} \in \mathbb{R}^{L \times K}$ is a weight and $\bm{b} \in \mathbb{R}^L$ is a bias term.
\textbf{BN-ReLU}$(K)$ combines a batch normalization and a ReLU activation: given an input $\bm{x} = \left[ x_1, \ldots, x_N \right]^\top \in \mathbb{R}^K$, its output $\bm{y} = \left[ y_1, \ldots, y_N \right]^\top \in \mathbb{R}^K$ is obtained as follows:
\begin{equation}
  y_i = \max \bigg( 0, \frac{x_i - \mu_i}{\sqrt{\sigma_i^2 + \varepsilon}}
    \cdot w_i + b_i \bigg) \quad (1 \le i \le K), \label{eq:bn-relu}
\end{equation}
where $\mu_i, \sigma_i$ are the mean and standard deviation, and $w_i, b_i$ denote the weight and bias.
\textbf{MaxPool}$(K)$ updates a global feature $\bm{\phi}(\mathcal{P}) = \left[ \phi_1, \ldots, \phi_K \right]^\top \in \mathbb{R}^K$ using a point-wise local feature $\bm{\psi}(\bm{p}) = \left[ \psi_1, \ldots, \psi_K \right]^\top \in \mathbb{R}^K$ as follows:
\begin{equation}
  \phi_i \gets \max(\phi_i, \psi_i) \quad (1 \le i \le K). \label{eq:max-pool}
\end{equation}

% where $\mu_i, \sigma_i$ are the $i$-th elements of the mean and standard deviation $\bm{\mu}, \bm{\sigma} \in \mathbb{R}^N$ estimated from the minibatches during training.
% Similarly, $w_i$ and $b_i$ are the $i$-th components of the weight and bias vectors $\bm{w}, \bm{b} \in \mathbb{R}^K$.
% The total number of parameters in \textbf{FC}$(K, L)$ and \textbf{BN-ReLU}$(K)$ are $KL + L$ and $4L$, respectively.
% In this paper, we exploit both intra- and inter-layer parallelism to minimize the latency, which are discussed in the following sections.

\subsection{Exploiting the Intra-layer Parallelism} \label{sec:design-intra-layer}
\textbf{FC}$(K, L)$ involves a matrix-vector multiplication $\bm{y} = \mathbf{W} \bm{x} + \bm{b}$ between a weight $\mathbf{W} \in \mathbb{R}^{L \times K}$ and an input $\bm{x} \in \mathbb{R}^K$, represented by two nested loops over $K$ and $L$.
We unroll the inner loop over $K$ by setting an unrolling factor to $B \ge 1$ to parallelize the multiplication between weights $w_{i, j}, \ldots, w_{i, j + B - 1}$ and inputs $x_j, \ldots, x_{j + B - 1}$.
The values $w_{i, k} x_k$ ($j \le k \le j + B - 1$) are then accumulated using an adder tree, which takes $\log B$ iterations.
In this way, the number of iterations is reduced from $KL$ to $L(K / B + \log B)$, which is roughly $B$x speedup.
This approach requires $B$ DSP blocks and the array partitioning of $\mathbf{W}, \bm{x}$ to increase the number of read operations per clock cycle.
We further reduce the latency by pipelining the inner loop.
% For instance, \textbf{FC}$(64, 64)$ modules for the second and third fully-connected layers (see Fig. \ref{fig:pointnet-layers}) are pipelined and parallelized by setting $B$ to 16, which results in the 16.3x speedup (latency is reduced from 8.39$\mu$s to 513ns).
\textbf{BN-ReLU}$(K)$ and \textbf{MaxPool}$(K)$ are easily parallelizable, as the computation for each output ($y_i$ or $\phi_i$) is independent as seen in Eqs. \ref{eq:bn-relu} and \ref{eq:max-pool}.
We set the unrolling factor $B \ge 1$ to compute multiple output elements ($y_i, \ldots, y_{i + B}$ or $\phi_i, \ldots, \phi_{i + B}$) and obtain $B$x performance improvement.

\subsection{Exploiting the Inter-layer Parallelism} \label{sec:design-inter-layer}
\begin{figure}[htbp]
  \centering
  \includegraphics[keepaspectratio,width=0.8\linewidth]{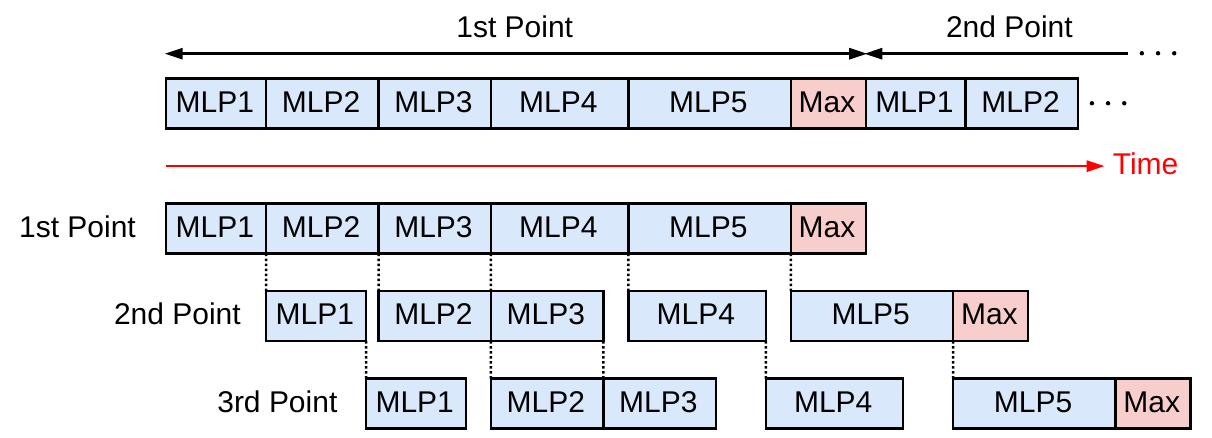}
  \vspace*{-2.5pt}
  \caption{Pipelined execution inside the PointNet IP core}
  \label{fig:dataflow-optimization}
\end{figure}

We also exploit the coarse-grained task-level parallelism to further improve the performance.
As depicted in Fig. \ref{fig:dataflow-optimization}, the modules work in a pipelined manner: this allows to overlap the computations for multiple input points and hide the data transfer overhead.
For instance, the fifth MLP layer (MLP5) computes a 1024D local feature of the first point, while the fourth MLP layer (MLP4) computes a 128D local feature of the second point.
We carefully choose a loop unrolling factor $B$ for each module to make the latency of all modules as even as possible (i.e., the pipeline evenly divides the workload among modules) and maximize the effectiveness of pipelining.
Table \ref{tbl:module-latencies} lists the unrolling factors and latencies ($B$ and $T$) for modules inside the core.
As expected, \textbf{FC}$(128, 1024)$ module for the last fully-connected layer is a bottleneck of the pipeline: we use the maximum possible value $B = 128$ to fully unroll the loop.
For the other modules, we adjust the unrolling factor $B$ such that their latencies do not exceed the one of \textbf{FC}$(128, 1024)$.

\begin{table}[h]
  \centering
  \caption{Unrolling factors and latencies for modules}
  \label{tbl:module-latencies}
  \vspace{-2.5pt}
  \begin{tabular}{l|rr|l|rr} \hline
    Module & $B$ & $T$ ($\mu$s) &
    Module & $B$ & $T$ ($\mu$s) \\ \hline
    \textbf{FC}$(3, 64)$ & 1 & 5.77 &
    \textbf{BN-ReLU}$(64)$ & 1 & 0.68 \\
    \textbf{FC}$(64, 64)$ & 16 & 5.13 &
    \textbf{BN-ReLU}$(128)$ & 1 & 1.32 \\
    \textbf{FC}$(64, 128)$ & 32 & 7.69 &
    \textbf{BN-ReLU}$(1024)$ & 2 & 5.16 \\
    \textbf{FC}$(128, 1024)$ & 128 & \textbf{10.28} &
    \textbf{MaxPool}$(1024)$ & 2 & 5.14 \\ \hline
  \end{tabular}
\end{table}

% \begin{table}[h]
%   \centering
%   \caption{Unrolling factors and latencies for modules}
%   \label{tbl:module-latencies}
%   \vspace{-2.5pt}
%   \begin{tabular}{l|rr} \hline
%     Layer & Unrolling factor $B$ & Latency ($\mu$s) \\ \hline
%     \textbf{FC}$(3, 64)$ & 1 & 5.77 \\
%     \textbf{FC}$(64, 64)$ & 16 & 5.13 \\
%     \textbf{FC}$(64, 128)$ & 32 & 7.69 \\
%     \textbf{FC}$(128, 1024)$ & 128 & \textbf{10.28} \\ \hline
%     \textbf{BN-ReLU}$(64)$ & 1 & 0.68 \\
%     \textbf{BN-ReLU}$(128)$ & 1 & 1.32 \\
%     \textbf{BN-ReLU}$(1024)$ & 2 & 5.16 \\ \hline
%     \textbf{MaxPool}$(1024)$ & 2 & 5.14 \\ \hline
%   \end{tabular}
% \end{table}

%% file: implementation.tex
% implementation.tex

\section{Implementation Details} \label{sec:implementation}
\begin{figure}[htbp]
  \centering
  \includegraphics[keepaspectratio,width=0.7\linewidth]{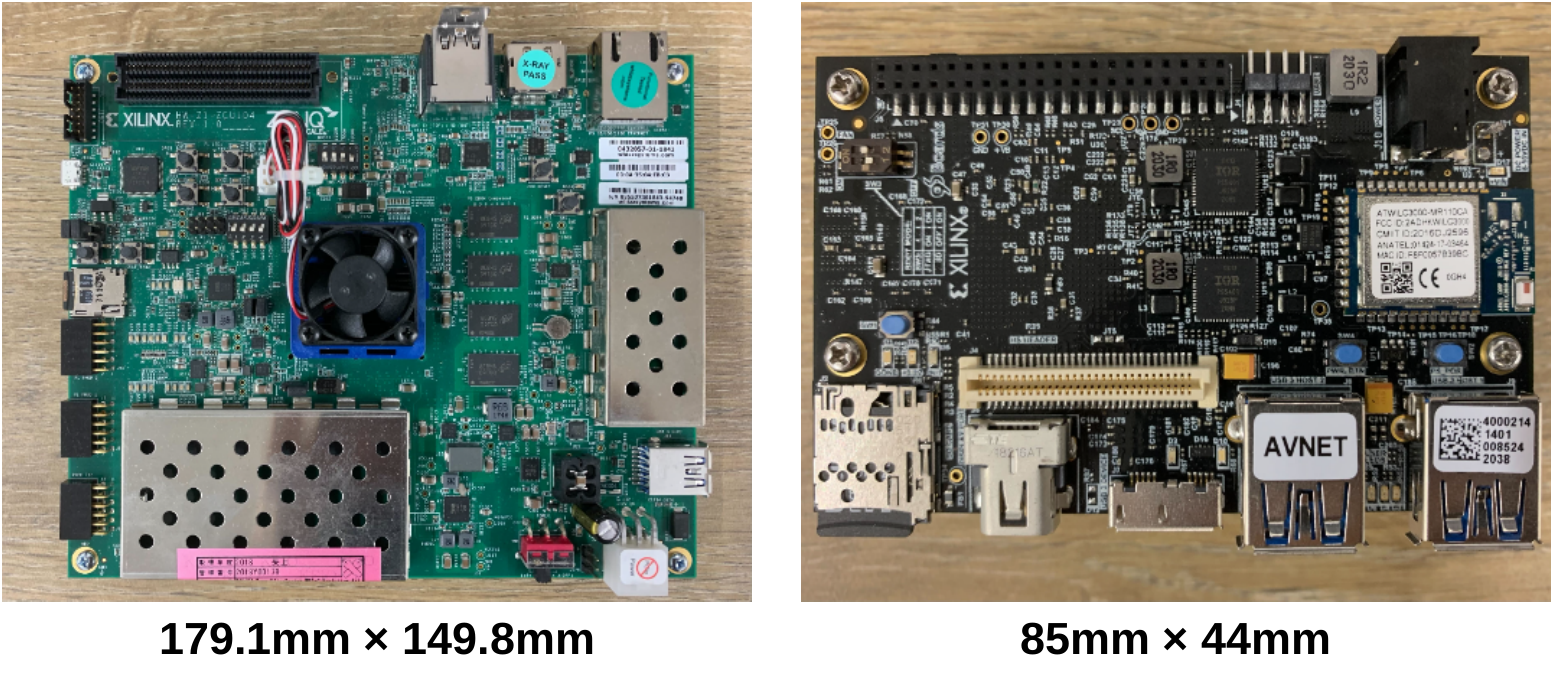}
  \vspace*{-5pt}
  \caption{FPGA boards (left: Xilinx ZCU104, right: Avnet Ultra96v2)}
  \label{fig:fpga-boards}
\end{figure}

We developed a custom accelerator for PointNet using Xilinx Vitis HLS 2020.2, and used Xilinx Vivado 2020.2 for synthesis and place-and-route.
We chose Xilinx Zynq UltraScale+ MPSoC devices, namely, Xilinx ZCU104 Evaluation Kit (XCZU7EV-2FFVC1156) and Avnet Ultra96v2 (ZU3EG A484) as target FPGA SoCs (Fig. \ref{fig:fpga-boards}), which integrates an FPGA and a mobile CPU on the same board.
The specifications of these chips are listed in Table \ref{tbl:fpga-spec}.
They both run Ubuntu 20.04-based Pynq Linux 2.7 on a quad-core ARM Cortex-A53 CPU at 1.2GHz and have a 2GB of DRAM.
We set the operation frequency of our accelerator to 100MHz.

% The PointNet IP core was written in C/C++, where each function describes the behavior of one module, i.e., \textbf{FC}, \textbf{BN-ReLU}, and \textbf{MaxPool}.
% We inserted preprocessor pragma directives provided by Vitis HLS to C/C++ source in order to perform necessary optimizations such as array partitioning, loop unrolling, inter-layer pipelining (\textit{dataflow optimization}), and high-speed DMA transfers, as described in Section \ref{sec:design}.
% They feature Xilinx Zynq UltraScale+ MPSoC chips; the former contains XCZU7EV-2FFVC1156, and the latter contains ZU3EG A484, respectively.
% They both run Ubuntu 20.04-based Pynq Linux 2.7 on a quad-core ARM Cortex-A53 CPU at 1.2GHz and have a 2GB of DRAM.
% We set the operation frequency of our accelerator to 100MHz.

\begin{table}[h]
  \centering
  \caption{FPGA specifications of Xilinx ZCU104 and Avnet Ultra96v2}
  \label{tbl:fpga-spec}
  \vspace{-2.5pt}
  \begin{tabular}{l|rrrr} \hline
    Board & BRAM & DSP & FF & LUT \\ \hline
    ZCU104 & 312 & 1728 & 460800 & 230400 \\
    Ultra96v2 & 216 & 360 & 141120 & 70560 \\ \hline
  \end{tabular}
\end{table}

We took the PointNetLK source code used in the original paper~\cite{Aoki19}, and modified it to offload PointNet feature extraction to our FPGA accelerator.
The code is implemented using Python 3.8.2 with PyTorch 1.10.2.
For ZCU104 and Ultra96v2, PyTorch was compiled using GCC 9.3.0 with ARM Neon intrinsics enabled to take advantage of the quad-core CPU.
The authors \cite{Aoki19} first pretrained the PointNet classification network and fine-tuned its weights by a PointNetLK loss function.
On the other hand, we trained PointNetLK from scratch and did not apply a transfer learning approach.
We used the same setting of hyperparameters as in the original code, and did not conduct a further parameter search.
The number of training epochs is set to 250.

%% file: eval.tex
\section{Evaluation} \label{sec:eval}
\subsection{Accuracy} \label{sec:eval-accuracy}
In this section, we evaluate the registration accuracy of PointNetLK using our proposed IP core in comparison with the CPU version and ICP~\cite{Besl92}.
As done in the original paper~\cite{Aoki19}, we trained PointNetLK on the training sets of 20 object classes (airplane to lamp) in ModelNet40~\cite{ZhirongWu15}, and tested on the test sets of the same 20 classes.

\begin{figure}[htbp]
  \centering
  \includegraphics[keepaspectratio,width=\linewidth]{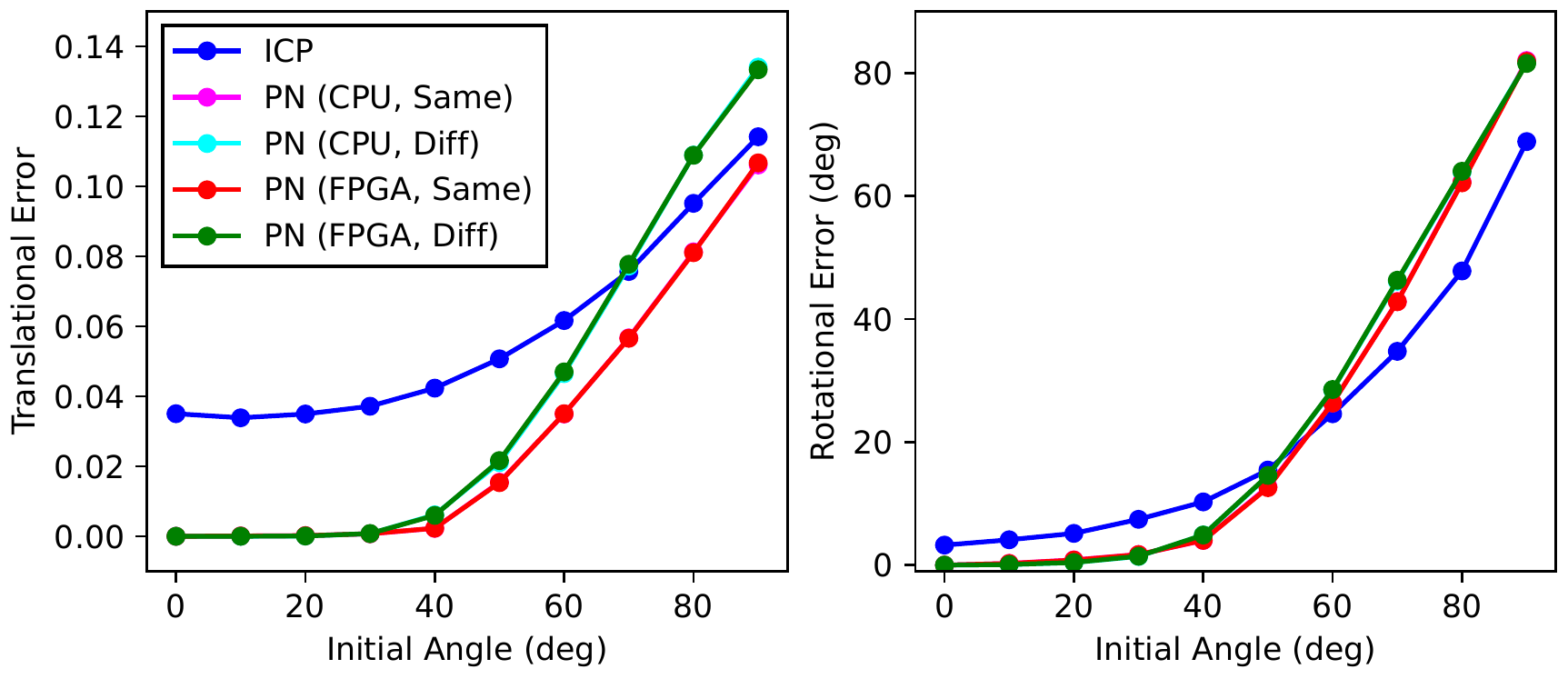}
  \vspace*{-17.5pt}
  \caption{Registration errors for PointNetLK and ICP}
  \label{fig:ex1-graph0}
\end{figure}

For each CAD model, we extracted a template point cloud $\mathcal{P}_T$ from the vertices, and normalized it to fit inside the unit cube.
We rotated $\mathcal{P}_T$ around a random axis by a constant angle $0^\circ \le \theta \le 90^\circ$, and then translated it by a random vector with each element uniformly distributed on $\left[ 0, 0.3 \right)$ to generate a source $\mathcal{P}_S$.
From a ground-truth transform $\mathbf{G}$ and an estimated transform $\mathbf{G}'$, we computed rotational and translational errors.
We downsampled (or upsampled) the input point clouds as necessary to fix the number of points $N$ in $\mathcal{P}_T$ and $\mathcal{P}_S$ to 1024 for all data samples.
In both ICP and PointNetLK, the same dataset $(\mathcal{P}_T, \mathcal{P}_S)$ and ground-truth $\mathbf{G}$ were used, and the maximum number of iterations was set to 20 for a fair comparison.
Fig. \ref{fig:ex1-graph0} shows the results with varying initial angles.

% To fix the number of points $N$ in $\mathcal{P}_T$ and $\mathcal{P}_S$ to 1024 for all data samples, we downsampled (or upsampled) the input point clouds as necessary.
% In both ICP and PointNetLK, the same dataset $(\mathcal{P}_T, \mathcal{P}_S)$ and ground-truth $\mathbf{G}$ were used, and the maximum number of iterations were set to 20 for a fair comparison.
% Fig. \ref{fig:ex1-graph0} shows the experimental results with different rotation angles $\theta$ between $\mathcal{P}_T$ and $\mathcal{P}_S$.

PointNetLK with our proposed IP core (magenta) achieves almost the same accuracy as the software implementation (red), and PointNetLK provides better accuracy than ICP for $\theta \le 50^\circ$.
For $\theta \ge 60^\circ$, PointNetLK does not converge to correct solutions and showed larger rotational errors than ICP.
This is an expected behavior; during training, we created a rigid transform $\exp \left( \bm{\xi}^\wedge \right) \in \SE$ between point clouds from a 6D vector $\bm{\xi}$ with norm less than 0.8.
In other words, PointNetLK was never trained on point cloud pairs with initial angles larger than 0.8 radians ($45.9^\circ$).

% The norm of an angle-axis rotation vector inside $\bm{\xi}$, i.e., the rotation angle between source $\mathcal{P}_S$ and template $\mathcal{P}_T$ is no more than 0.8 radians (45.9 degrees), meaning that the PointNetLK was never trained on pairs of point clouds with such large orientation differences.

\begin{figure}[htbp]
  \centering
  \includegraphics[keepaspectratio,width=0.65\linewidth]{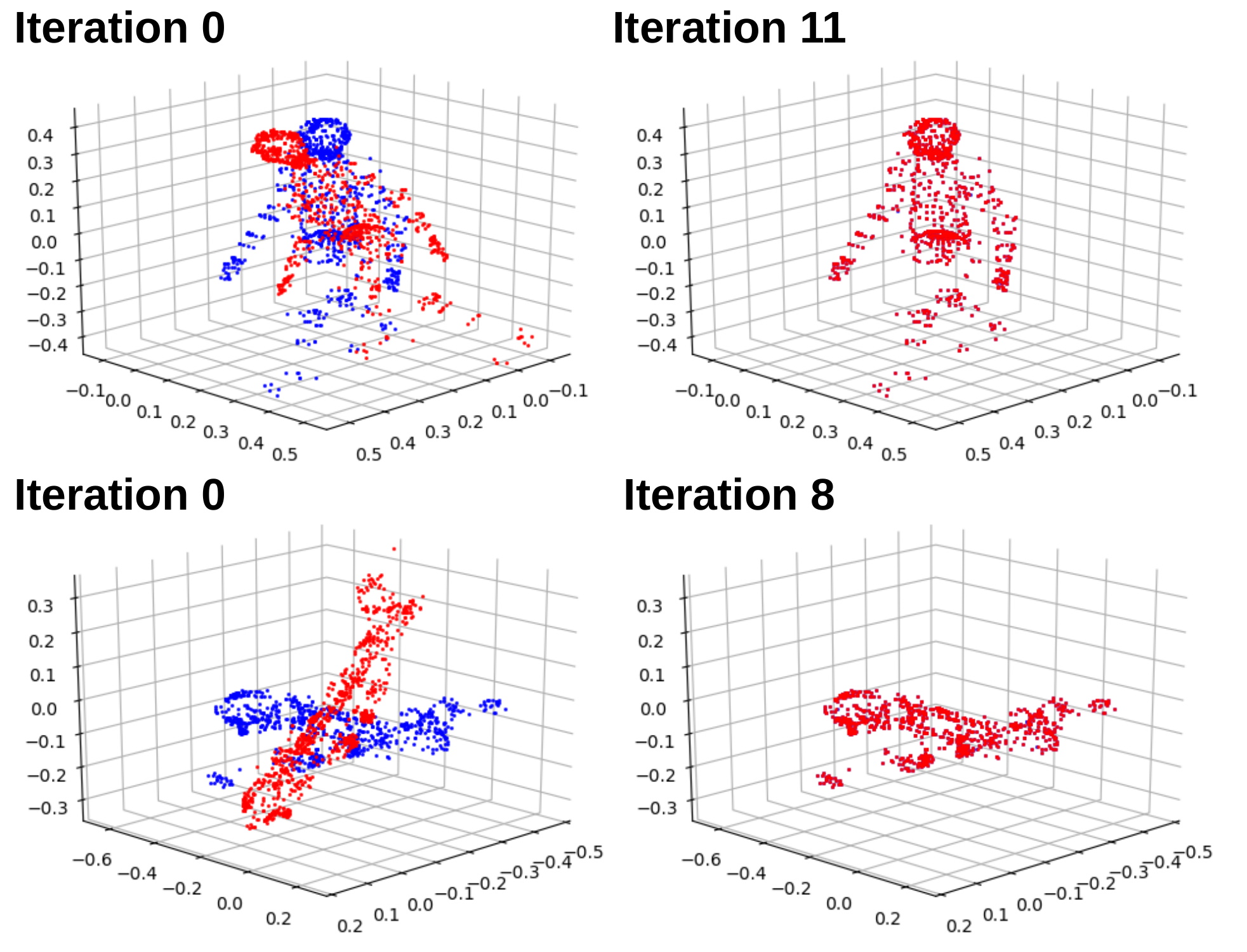}
  \vspace*{-2.5pt}
  \caption{Examples of the registration using PointNetLK (ModelNet40).
  To show the generalization ability, PointNetLK trained on the first 20 object classes (airplane to lamp) was employed for the person model (top), whereas the one trained on the last 20 categories (laptop to xbox) was used for the airplane model (bottom).
  Red and blue points represent the source and template, respectively.
  The initial angle was set to $60^\circ$.}
  \label{fig:registration-examples}
\end{figure}

\begin{figure}[htbp]
  \centering
  \includegraphics[keepaspectratio,width=0.65\linewidth]{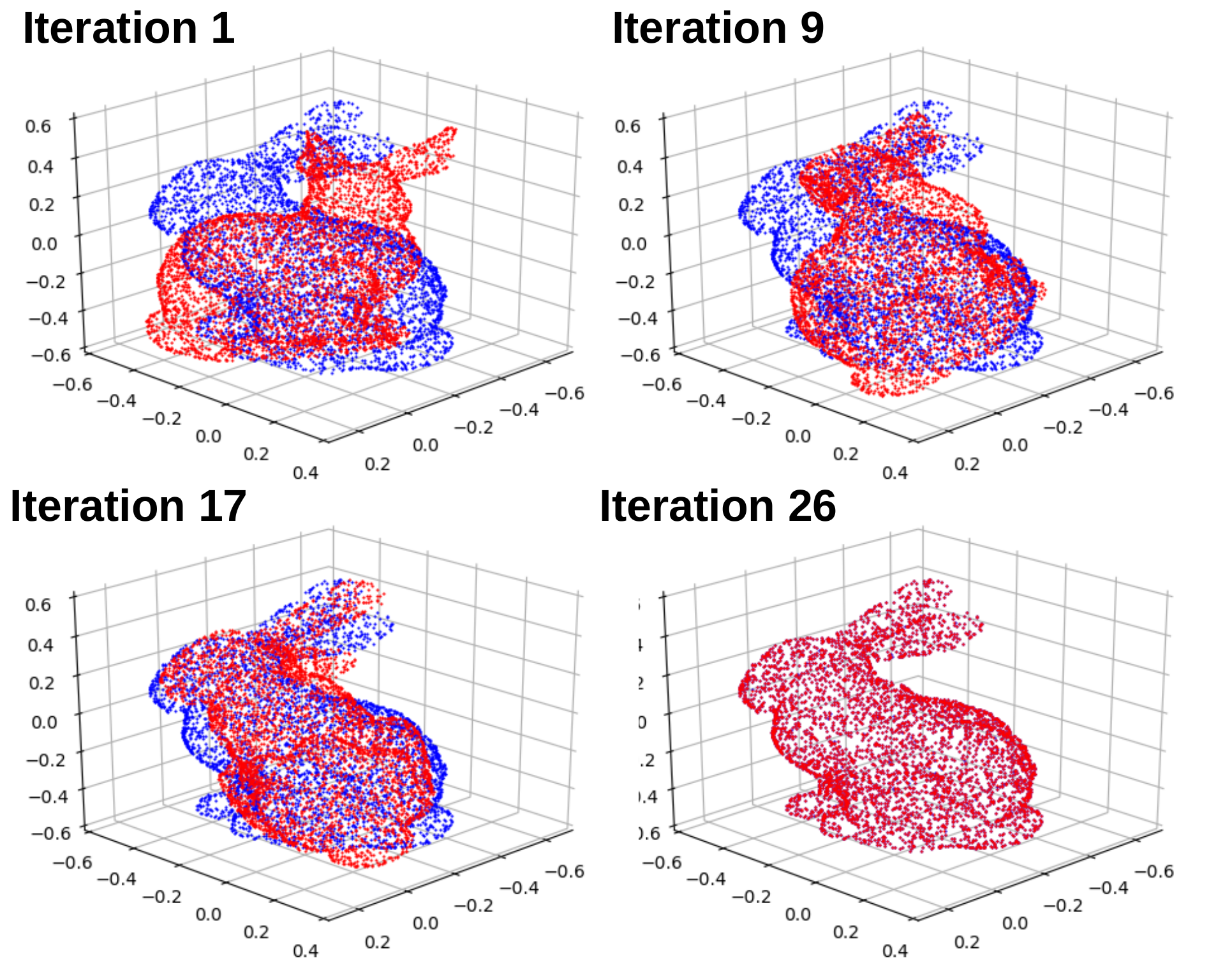}
  \vspace*{-2.5pt}
  \caption{Examples of the registration using PointNetLK (Stanford bunny).
  PointNetLK trained on the 20 categories (airplane to lamp) in ModelNet40 successfully aligns a pair of bunny models which is distinct from the training set, thereby showing a generalization ability.
  A template (blue points) was rotated $90^\circ$ around a random axis to generate a source (red points).}
  \label{fig:bunny-registration}
\end{figure}

We also trained PointNetLK on the training sets of 20 classes and tested on the test sets of the other 20 classes (laptop to xbox).
While it (cyan, green) shows a larger translational error than PointNetLK trained and tested with the same classes (magenta, red) for $\theta \ge 60^\circ$, it still achieves the same level of accuracy especially in the rotation estimation.
Besides, for $\theta \le 50^\circ$, the registration error is lower than ICP and closer to that of PointNetLK in the previous setting.
This indicates that PointNetLK has a generalization ability to align point clouds which are distinct from the training dataset.
As shown in Fig. \ref{fig:ex1-graph0}, PointNetLK with the FPGA acceleration (green) has almost the same accuracy as the software counterpart (cyan), meaning that our IP core yields faster computation time without compromising the accuracy.
For qualitative analysis, Figs. \ref{fig:registration-examples} and \ref{fig:bunny-registration} visualize the registration results obtained from PointNetLK with our IP core for ModelNet40 and Stanford bunny~\cite{Turk05}, respectively.

% We also trained PointNetLK on the training sets of 20 classes (airplane to lamp) and tested on the test sets of the other 20 classes (laptop to xbox) containing 1266 point clouds.

% For ease of visualization, we translate both source and template so that their centroids lie at the origin.

\subsection{Computation Time} \label{sec:eval-computation-time}
\begin{figure}[htbp]
  \centering
  \includegraphics[keepaspectratio,width=0.65\linewidth]{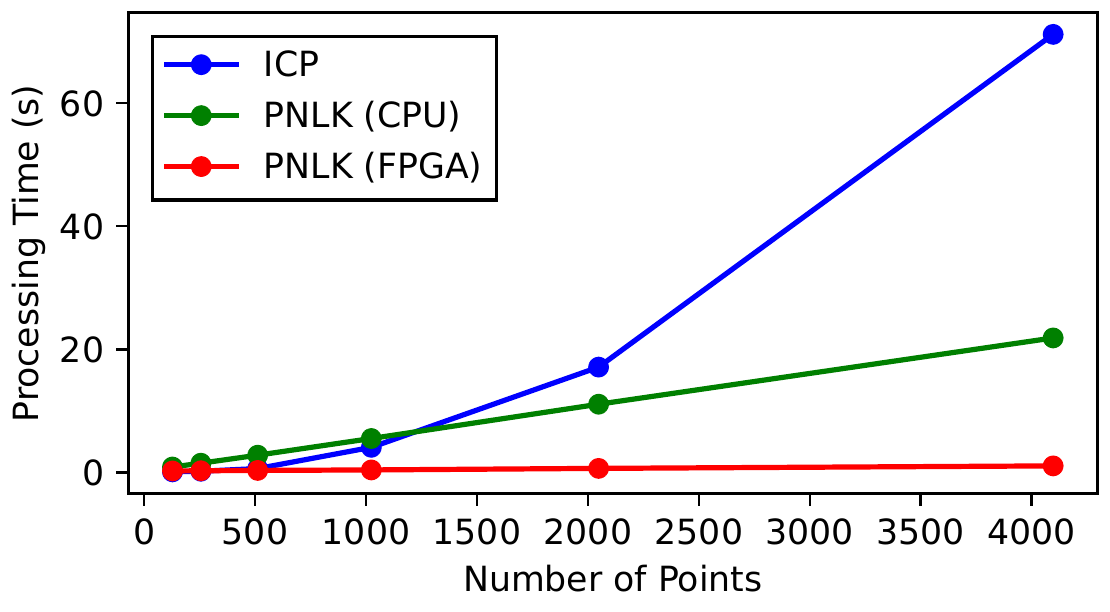}
  \vspace*{-5pt}
  \caption{Processing time for PointNetLK and ICP}
  \label{fig:ex2-graph0}
\end{figure}

PointNetLK is evaluated in terms of the computation time to highlight its significant advantage over ICP.
Fig. \ref{fig:ex2-graph0} shows the results with the varying number of input points $N$ from 128 to 4096.
We used the table category in ModelNet40 and plotted an average wall-clock time.
We included the data transfer overhead between PS--PL for a fair comparison.
The initial angle $\theta$ is set to $30^\circ$.
We also note that PointNetLK was trained on the first 20 categories in ModelNet40, which do not include the table category.
The wall-clock time increases linearly in PointNetLK and quadratically in ICP, which stems from the fact that the computational complexity of PointNetLK and ICP are $O(N)$ and $O(N^2)$, respectively.
It directly follows that PointNetLK provides a better performance advantage over ICP as the input size $N$ increases.
For $N = 1024$, the CPU version of PointNetLK was 1.36x slower than ICP (5.47s and 4.04s).
The FPGA version (red) took only 366ms per input, which was 14.98x faster compared to the CPU version (green), and eventually lead to 11.04x speedup than ICP (blue).
As shown in Fig. \ref{fig:ex2-graph0}, we obtained better results for $N = 4096$: compared to ICP, the CPU version was 3.26x faster (71.16s and 21.82s), and the FPGA version was 69.60x faster, which is attributed to the 21.34x speedup (21.82s to 1.02s).

% We used the table category in ModelNet40 as inputs, which contains a set of 100 point clouds, and plotted an average wall-clock time for registration.
% We followed the experiment in \cite{Aoki19} and did not employ a $k$d-tree in the ICP implementation, meaning that it involves a nearest neighbor search with $O(N^2)$ complexity to find point correspondences between template and source clouds.

\begin{figure}[htbp]
  \centering
  \includegraphics[keepaspectratio,width=0.85\linewidth]{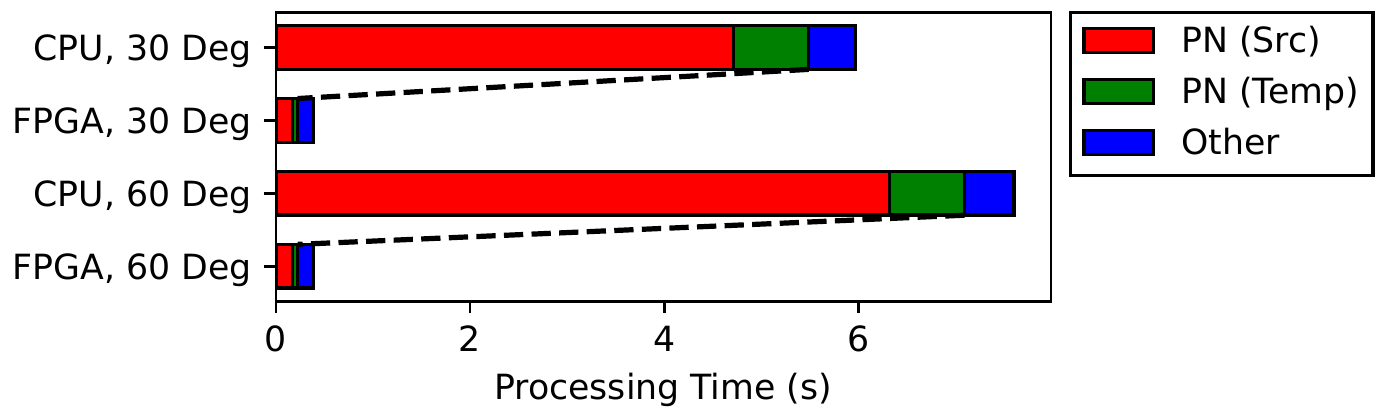}
  \vspace*{-5pt}
  \caption{Breakdown of the processing time}
  \label{fig:ex5-graph0}
\end{figure}

Fig. \ref{fig:ex5-graph0} shows the breakdown of processing time for PointNetLK with and without FPGA acceleration.
We set the initial angles $\theta$ to $30^\circ$ (first two rows) and $60^\circ$ (last two rows).
PointNet feature extraction (red + green) is inevitably a major bottleneck, accounting for 91.90\% ($\theta = 30^\circ$) and 93.29\% ($\theta = 60^\circ$), which were reduced to 58.01\% and 57.96\% by FPGA acceleration.

% We again measured the processing time using the table category in ModelNet40, and set the initial rotation $\theta$ to $30^\circ$ (first two rows) and $60^\circ$ (last two rows).
% PointNet feature extraction (red + green) is inevitably a major bottleneck, accounting for 91.90\% and 93.29\% of PointNetLK for $\theta = 30^\circ$ and $60^\circ$, respectively.
% After the FPGA acceleration, they were reduced to 58.01\% and 57.96\%, demonstrating the effectiveness of our proposal.
% In the CPU version (1st and 3rd rows), the PointNet inference time for source (red) increased by a noticeable margin, since it needed more iterations to converge to a final solution (13.23 and 17.73 iterations on average for $\theta = 30^\circ, 60^\circ$).
% Despite that PointNetLK with our accelerator (2nd and 4th rows) took nearly 20 iterations to reach convergence for both settings, which is the predefined maximum in this experiment, it achieved almost the same accuracy as shown in Fig. \ref{fig:ex1-graph0}.

\subsection{Effects of Quantization} \label{sec:eval-quantization-effects}
\begin{figure}[htbp]
  \centering
  \includegraphics[keepaspectratio,width=\linewidth]{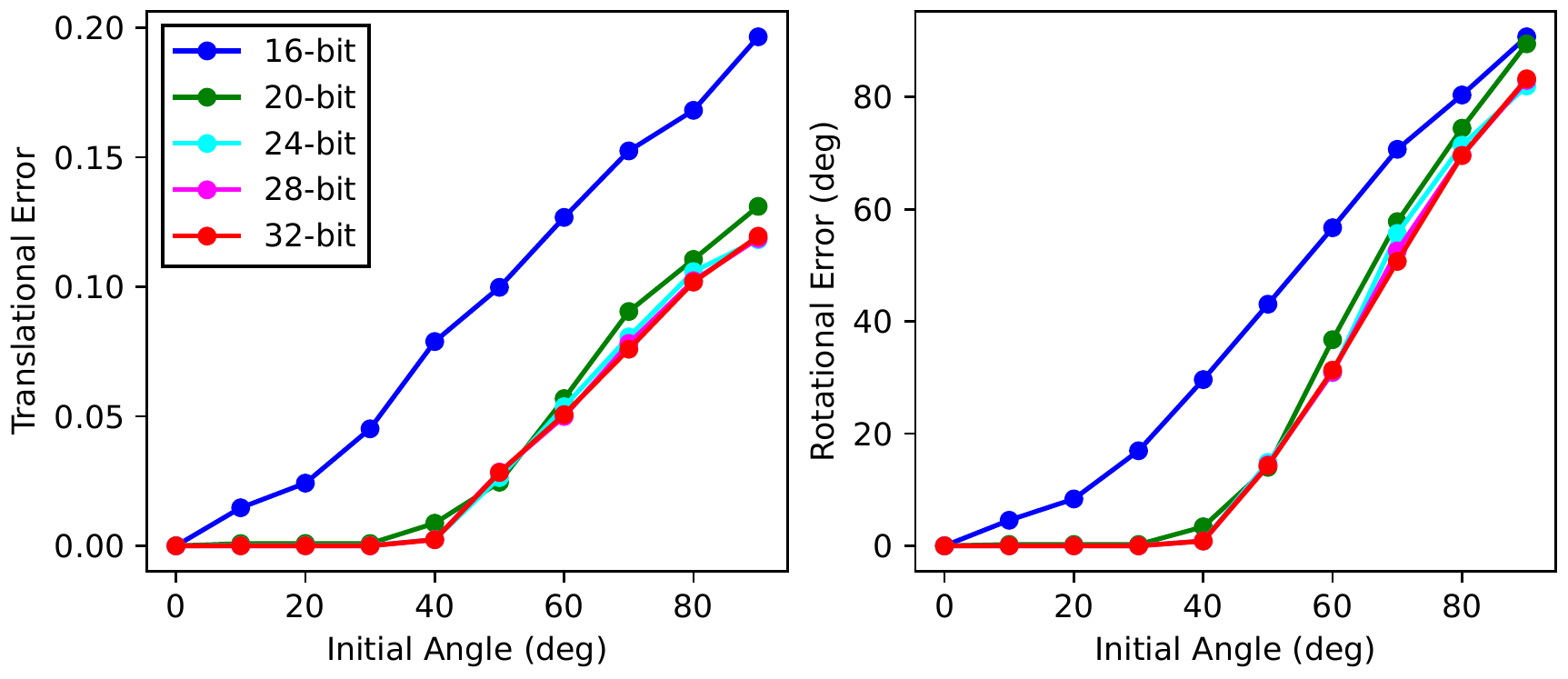}
  \vspace*{-17.5pt}
  \caption{Registration errors and the number of quantization bits}
  \label{fig:ex4-graph0}
\end{figure}

This section analyzes a relationship between the number of quantization bits used in the IP core and the accuracy.
Fig. \ref{fig:ex4-graph0} shows the PointNetLK registration errors, evaluated using five different numbers of quantization bits from 16 to 32.
Table \ref{tbl:resource-utilization-quantization} summarizes the FPGA resource utilization.
Each IP core design uses the $2n$-bit fixed-point format with $n$-bit integer part and $(n - 1)$-bit fraction part ($n = 8, 10, 12, 14, 16$).
We trained PointNetLK on the first 20 object classes and tested with the table class in ModelNet40.

% In the experiment, the number of input points $N$ is set to 1024.

As apparent in Fig. \ref{fig:ex4-graph0}, the 16-bit quantized version exhibited larger errors than the others.
In contrast, for $0^\circ \le \theta \le 50^\circ$, the 20-bit version produced nearly the same results as the 32-bit version.
Even for $\theta \ge 60^\circ$, the bitwidth reduction from 32 to 20-bit only introduced a slight accuracy loss.
Notably, the DSP usage was halved by reducing from 32 to 24-bit (Table \ref{tbl:resource-utilization-quantization}).
The reduction from 24 to 20-bit further halved the DSP footprint (24.07\% to 12.56\%) and increased the LUT usage (9.91\% to 12.87\%), since arithmetic units such as multipliers were implemented using more LUTs and less DSPs.
The results indicate that the 20-bit version strikes the best balance between accuracy and resource consumption.
As seen in Table \ref{tbl:resource-utilization-ultra96v2}, the 20-bit version fits within a low-cost and resource-limited FPGA, Avnet Ultra96v2, whereas the 32-bit version cannot be implemented due to the shortage of DSP and LUT resources.

% The Vivado toolchain tries to use LUTs as memory to compensate for the BRAM shortage, which leads to the excessive LUT usage.

% As apparent in Fig. \ref{fig:ex4-graph0}, the design with 16-bit quantized PointNet exhibited larger registration errors than the others.
% In contrast, for initial rotations from $0^\circ$ to $50^\circ$, the design with 20-bit quantization produced nearly the same results as the one with 32-bit quantization.
% Even for initial rotations larger than $60^\circ$, the bitwidth reduction from 32-bit to 20-bit only introduced a slight accuracy loss.

\begin{table}[h]
  \centering
  \caption{FPGA resource utilization and quantization}
  \label{tbl:resource-utilization-quantization}
  \vspace{-2.5pt}
  \begin{tabular}{r|rrrr} \hline
    \# of Bits & BRAM (\%) & DSP (\%) & FF (\%) & LUT (\%) \\ \hline
    32 & 55.13 & 48.50 & 5.46 & 16.31 \\
    % 172 / 312, 838 / 1728, 25156 / 460800, 37579 / 230400
    28 & 55.13 & 48.09 & 4.73 & 13.70 \\
    % 172 / 312, 831 / 1728, 21773 / 460800, 31574 / 230400
    24 & 44.87 & 24.07 & 4.05 & 9.91 \\
    % 140 / 312, 416 / 1728, 18657 / 460800, 22839 / 230400
    20 & 44.23 & 12.56 & 3.85 & 12.87 \\
    % 138 / 312, 217 / 1728, 17723 / 460800, 29660 / 230400
    16 & 27.40 & 12.21 & 2.83 & 7.74 \\ \hline
    % 85.5 / 312, 211 / 1728, 13054 / 460800, 17822 / 230400
  \end{tabular}
\end{table}

\begin{table}[h]
  \centering
  \caption{FPGA resource utilization on Avnet Ultra96v2}
  \label{tbl:resource-utilization-ultra96v2}
  \vspace{-2.5pt}
  \begin{tabular}{l|rrrr} \hline
    \# of Bits & BRAM (\%) & DSP (\%) & FF (\%) & LUT (\%) \\ \hline
    20 & 64.81 & 60.28 & 12.81 & 43.52 \\
    % 140 / 216, 217 / 360, 18074 / 141120, 30709 / 70560
    32 & 79.63 & 100.00 & 22.35 & \textbf{238.77} \\ \hline
    % 172 / 216, 360 / 360, 31535 / 141120, 168478 / 70560
  \end{tabular}
\end{table}

\subsection{Effects of Design Optimization} \label{sec:eval-design-optimization-effects}
\begin{figure}[htbp]
  \centering
  \includegraphics[keepaspectratio,width=0.7\linewidth]{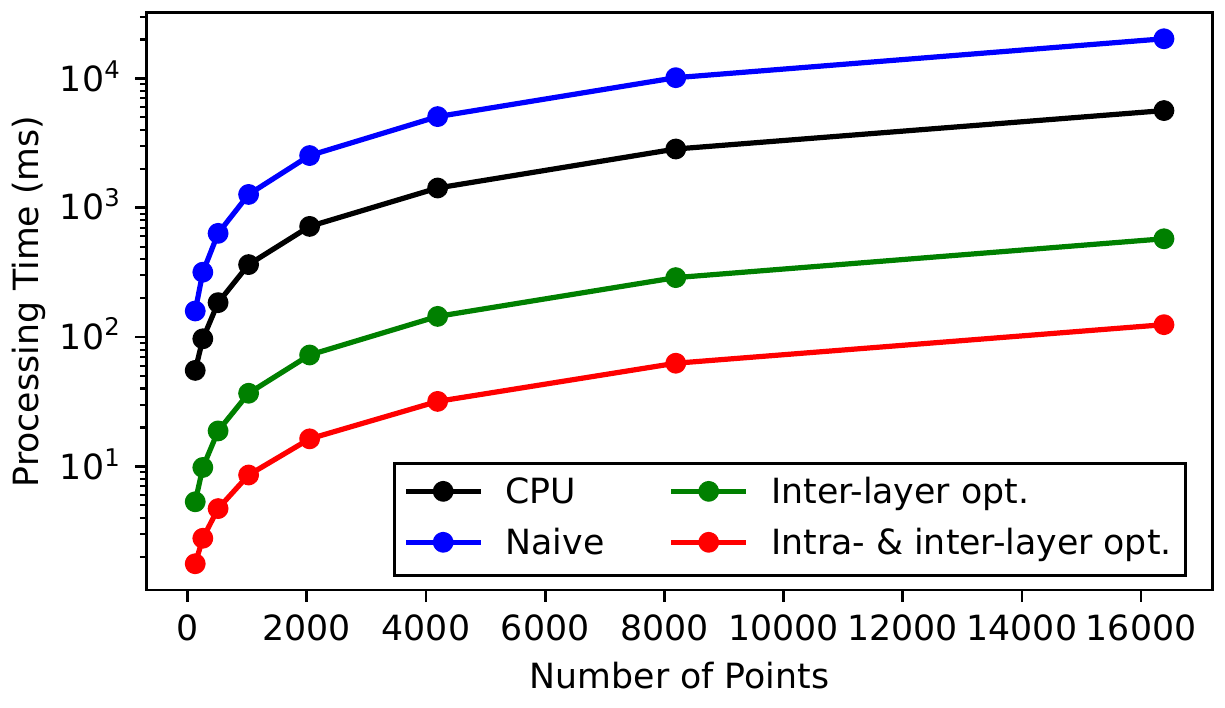}
  \vspace*{-5pt}
  \caption{Comparison of inference time of PointNet}
  \label{fig:ex8-graph}
\end{figure}

Here, we discuss the effects of design optimizations described in Section \ref{sec:design} on the performance and FPGA resource utilization.
In addition to the final design, we also consider the design without inter-layer pipelining and the naive design with no optimization as a baseline.
Fig. \ref{fig:ex8-graph} plots the processing time with varying point cloud sizes from $N = 128$ to $N = 16384$, and Table \ref{tbl:resource-utilization} compares the resource utilization.
In Fig. \ref{fig:ex8-graph}, we observe a linear increase of the processing time, and the naive design (blue) is 3.49x slower than the CPU (black) for $N = 1024$ (363.49ms and 1267.08ms).
By exploiting the intra-layer parallelism, the design (green) attains a speedup of 34.46x ($N = 1024$) compared to the unoptimized version (1267.08ms to 36.77ms) at the expense of 14.38x increase in the DSP usage (3.07\% to 44.16\%).
The intra-layer pipelining allows a further speedup of 4.29x (green and blue, 36.77ms to 8.58ms) with a few additional resources, by overlapping data transfer and computation.
This leads to a total performance improvement of 147.68x and 42.36x compared to the unoptimized version and CPU for $N = 1024$, respectively.

% In addition to the PointNet IP core design exploiting intra- and inter-layer parallelism, we also consider the design without inter-layer pipelining and the naive design with no optimization as a baseline.

\begin{table}[h]
  \centering
  \caption{FPGA resource utilization and design optimization}
  \label{tbl:resource-utilization}
  \vspace{-2.5pt}
  \begin{tabular}{l|rrrr} \hline
    Design & BRAM (\%) & DSP (\%) & FF (\%) & LUT (\%) \\ \hline
    Naive & 45.99 & 3.07 & 0.82 & 3.73 \\
    % 143.5 / 312, 53 / 1728, 3783 / 460800, 8598 / 230400
    Intra-layer & 54.17 & 44.16 & 4.56 & 10.66 \\
    % 169 / 312, 763 / 1728, 21012 / 460800, 24555 / 230400
    Inter- \& intra-layer & 55.13 & 48.50 & 5.46 & 16.31 \\ \hline
    % 172 / 312, 838 / 1728, 25156 / 460800, 37579 / 230400
  \end{tabular}
\end{table}

\subsection{Power Consumption} \label{sec:eval-power-consumption}
% The power consumption of the entire ZCU104 Evaluation board was 13.4W when running PointNetLK on the CPU, and reached 15.7W when running it with our accelerator.
% Note that these values include the power consumption of CPU and FPGA fabric, as well as that of the other peripherals; the accelerator itself consumes much less power than the above.
The power consumption of our accelerator was 722mW according to the estimates reported by Xilinx Vivado 2020.2.

% We present the power consumption of the entire ZCU104 evaluation board measured by a wattmeter.
% In the idle state, the power consumption was around 13.0W; it increased to 13.4W when running PointNetLK on the CPU, and reached 15.7W when running PointNetLK in conjunction with our proposed FPGA accelerator.
% We therefore present the estimates reported by Vivado 2020.2 after the synthesis and place-and-route steps.
% The power consumption of our accelerator was 722mW, accounting for 20\% of the total power consumption of CPU and FPGA fabric.

%% file: conc.tex
% conc.tex

\section{Conclusion} \label{sec:conc}
In this paper, we present a resource-efficient FPGA-based implementation for 3D point cloud registration.
We opt to use PointNetLK, which combines PointNet feature embedding and Lucas-Kanade (LK)-based pose estimation.
We develop a custom PointNet accelerator and implement it on a mid-range FPGA (Xilinx ZCU104).
We exploit both intra- and inter-layer parallelism in PointNet to fully optimize the design, achieving $O(N)$ computational complexity and $O(1)$ memory requirement.
Experiments demonstrate that PointNetLK with our accelerator achieves up to 21.34x and 69.60x speedup compared to the CPU counterpart and ICP, respectively, without compromising the accuracy.
Besides, it consumes only 722mW at runtime and offers better scalability than ICP.
The quantized design fits within even smaller FPGAs (Avnet Ultra96v2).
Experiments also highlight the generalization ability of PointNetLK.
% Experiments also show that PointNetLK is able to correctly align point clouds which are unseen during training, highlighting its generalization ability.